# *A la croisée de l'informatique et du théâtre : similarité en* **intension** *versus similarité en* **extension**

**Alain Bonardi**
Equipe Intelligence Artificielle et Robotique Mobile
Université Paris 8
2, rue de La Liberté, 93526 Saint-Denis
alain.bonardi@wanadoo.fr

**Francis Rousseaux**
Equipe projet SemanticHIFI de l'IRCAM
et CReSTIC de l'Université de Reims
1, place Igor-Stravinsky, 75004 Paris
francis.rousseaux@ircam.fr

**RÉSUMÉ**

La mise en scène de théâtre traditionnelle repose sur une approche formelle de la similarité s'appuyant sur des ontologies dramaturgiques et des variations d'instanciation. Inspirés par la fouille de données numériques interactive, qui suggère des approches différentes, nous rendons compte de recherches théâtrales utilisant l'ordinateur comme partenaire de l'acteur pour échapper à la spécification *a priori* des rôles.

**MOTS-CLÉS**

similarité, instanciation, ontologies, fouille de données interactive, théâtre inter-média.

## 1. INTRODUCTION

Dans cet article, nous réfléchissons à la mise en scène de théâtre et ses évolutions dans le contexte du dialogue acteurs-ordinateurs, en mobilisant les catégories de l'informatique. Nous avons déjà abordé l'approche inverse en montrant comment des progiciels tels que Powerpoint™ sont fondés sur des conceptions théâtrales et plus particulièrement scénographiques, liées aux notions d'avant-plan et d'arrière-plan [ROUSSEAUX & BONARDI02].

Distinguant deux approches de la similarité en informatique – inspirées par les ontologies *vs* par la fouille de données interactive, nous les utilisons pour comprendre la mise en scène au sens traditionnel et ses nouvelles modalités liées à l'utilisation d'ordinateurs comme partenaires des acteurs. Notre recherche s'appuie sur un exemple de réalisation, la pièce de théâtre inter-média *La traversée de la nuit* de Geneviève de Gaulle.

## 2. L'APPROCHE TRADITIONNELLE DE LA MISE EN SCÈNE DANS UNE PERSPECTIVE INFORMATIQUE

Face à un texte de théâtre, chaque metteur en scène souhaite proposer sa lecture/interprétation. Rappelons en effet que l'interprétation n'est pas immanente au texte, malgré les indications parfois nombreuses (préface, didascalies) de l'auteur. Un texte ne peut exister sur scène sans exégèse du metteur en scène.

## 2.1. Ontologies de la dramaturgie et variations d'instanciation

Cette lecture/interprétation est toujours un effort pour créer des formes. Essayons d'en rendre compte dans une perspective informatique. La démarche du metteur en scène commence par l'établissement d'une ontologie synthétique de la dramaturgie : on y décrit les personnages sous forme de types (ce en quoi le théâtre de boulevard par exemple excelle ave son trio, mari, femme et amant !) et d'instanciations, en indiquant le nom du personnage, sa situation au début de la pièce, son costume. Le déroulement de la pièce propose des variations d'instanciation[1] : le spectateur découvre que tel ou tel personnage est différent de ce qu'il imaginait au départ. Ces variations d'instanciation peuvent parfois aboutir à des révisions d'ontologie. C'est par exemple l'enjeu, autant métaphysique que théâtral, de la pièce *El burlador de Sevilla* du dramaturge espagnol Tirso de Molina (1630) qui inaugure le mythe de Don Juan : ce personnage peut-il être sauvé en reconnaissant ses fautes *in extremis* avant sa mort ? L'ontologie du personnage peut-elle être radicalement modifiée à la fin de la pièce ?

## 2.2. Une approche formelle de la similarité à base d'ontologies

Dans cette approche traditionnelle du théâtre, la notion de similarité par les ontologies est centrale. Le metteur en scène règle chaque scène ou passage faisant unité en le considérant comme un exemple dans un ensemble de cas fournis par la littérature théâtrale. Expliquer un personnage à l'acteur qui le joue revient à le pointer dans l'ontologie proposée et à relier cette ontologie à celle d'autres pièces ou d'autres lectures de la même pièce par d'autres metteurs en scène, pour donner à comprendre par un exemple dit « similaire ».

En généralisant, il s'agit d'une **approche formelle**, dans laquelle on représente l'exemple comme une instance d'une structure générale embrassant tous les cas, et on cherche les similarités en faisant varier l'instanciation. Cette approche présente l'avantage de fournir une explication du caractère « similaire à l'exemple » de la proposition, voire une mesure de distance: c'est par ce biais qu'un concept récapitulatif en *intension* peut être créé. Les ontologies permettent de rechercher les similarités à un exemple en demeurant dans l'enceinte du concept, quitte à passer au concept immédiatement plus général quand la quête est infructueuse. Ceci s'applique à bien d'autres activités que le théâtre, par exemple l'organisation de la vente de disques compacts (CD) dans un grand magasin de disques [ROUSSEAUX & BONARDI04b]. En effet, la pratique de l'achat de CD prescrit subrepticement nos activités musicales et le rangement *a priori* dans des bacs de vente, à évolution lente, est structuré par l'acquisition marchande et la notion de genre.

## 3. L'APPROCHE DE LA SIMILARITÉ PAR LA FOUILLE DE DONNÉES INTERACTIVE

Au niveau informatique, il existe une autre manière d'aborder la question de la similarité : c'est **l'approche fouille de données interactive**, dans laquelle on représente l'exemple comme une spécialisation de l'ensemble des cas, et on cherche d'autres spécialisations voisines, mais sans disposer par avance d'une ontologie. L'utilisateur accepte de la façonner à sa main avec l'aide interactive de la machine, de manière *ad hoc*. Il s'agit d'une approche **en extension** : façonner une similarité revient à façonner une liste de contenus de forme similaire par des opérations rectificatives successives mobilisant le calcul numérique en interaction interprétative avec des actions rectificatrices sur les contenus et leur forme (du côté de l'utilisateur, provoqué par les propositions de la machine).

L'activité de music-ripping illustre cette approche [ROUSSEAUX & BONARDI04]. Elle consiste en la manipulation créative de contenus audionumériques passant par des gestes de modification, aboutement, suppression, etc. associés aux interfaces informatiques. Lorsque l'activité pratiquée est une écoute signée (Donin, 2004), une écoute/composition/production, son objet devient le grain élémentaire d'écoute/composition/production, un échantillon, constamment modifié, réorganisé, re-mixé et renommé (Pachet, 2003) par l'utilisateur.

---

[1] *Instanciation* est un anglicisme couramment utilisé par les informaticiens, qui renvoie au mot *instance* signifiant *exemple, cas*. L'instanciation généralise en quelque sorte l'opération, utilisée par les mathématiciens, d'affectation d'une valeur numérique à une variable : pour parler du réel, les informaticiens instancient des classes abstraites, décrétant ainsi que telle ou telle entité est un cas particulier d'une classe, elle-même reliée à d'autres classes par des hiérarchies de généralité et/ou des propriétés formelles, l'ensemble du dispositif [PERROT94] constituant ce qu'on appelle parfois une *ontologie* (les ontologies prétendent ainsi décrire des pans de connaissances mondaines très utilisées en intelligence artificielle), parfois une *conception à objets* (une conception à objets est constituée de graphes d'héritage conçus pour donner lieu à des programmes informatiques par simple instanciation de paramètres clés).

Remarquons que nous touchons là à une des différences fondamentales entre les mathématiques et l'intelligence artificielle. En effet, les mathématiques posent l'équivalence entre *l'intension* et *l'extension*. La notion de classe d'équivalence est par exemple présente sur les deux versants : d'un côté, on peut relier deux éléments individuels entre eux en vérifiant qu'ils appartiennent ou pas à une même classe ; de l'autre, on recouvre les ensembles (par exemple l'ensemble des entiers relatifs) avec un nombre minimal de classes. Les noyaux d'endomorphismes jouent le même rôle en algèbre vectorielle : on peut vérifier que deux vecteurs individuels appartiennent ou non au même noyau et on peut dans le cas d'endomorphismes diagonalisables recouvrir un espace vectoriel par un nombre fini de noyaux. En revanche, dans le domaine de l'intelligence artificielle, il n'y a pas d'équivalence entre *intension* et *extension*, sans doute parce que l'équivalence entre un spécimen (comme instance particulière d'une catégorie) et une singularité (perçue comme traversant le réel) n'est tout au plus acceptable que comme mauvaise heuristique (car niant la notion même de situation).

## 4. RETOUR AU THÉÂTRE : LA *TRAVERSÉE DE LA NUIT*

Que peut donner cette approche informatique de fouille de données interactive au niveau de la mise en scène de théâtre ? Elle suppose l'introduction de l'ordinateur selon un mode de dialogue entre acteurs et machines. Une mise en scène peut-elle ne plus se conformer à une ontologie préexistante mais échapper à la spécification *a priori* en s'appuyant sur des interactions multimodales ? C'est le sens de la recherche que nous avons menée dans le spectacle de théâtre inter-média *La traversée de la nuit*[2], sur le texte de Geneviève de Gaulle-Anthonioz [DE GAULLE98], évoquant son emprisonnement au cachot du camp de Ravensbrück à la fin de la Deuxième Guerre Mondiale.

### 4.1. Les interactions multi-modales dans la *Traversée de la nuit*

La mise en scène de *La traversée de la nuit*, repose sur un système homme-machine « autarcique » : une comédienne, Valérie Le Louédec, disant l'intégralité du texte, une danseuse, Magali Bruneau, accomplissant un certain nombre de gestes inspirés du théâtre Nô et un ordinateur multimédia, acteur artificiel. L'ordinateur se manifeste sous forme d'images projetées sur un écran de fond de scène de très vastes dimensions (la comédienne et la danseuse en voient toujours au moins une partie sans se retourner), provoquant la réaction des deux comédiennes, notamment de la danseuse adaptant la réalisation de sa gestuelle aux mouvements et qualités de l'image. Or, les deux actrices sur scène constituent les deux versants du même personnage – conscient et inconscient, selon les traditions du *shite* et du *waki* du théâtre Nô. Entraînée dans ses déplacements par la danseuse, la comédienne adapte elle aussi sa déclamation, sans compter les moments où elle regarde aussi l'écran. Pour boucler la boucle, l'ordinateur capte les états émotionnels de la voix de la comédienne.

---

[2] Pièce de théâtre donnée les 21, 22 et 23 novembre 2003 au Centre des Arts d'Enghien-les-Bains (95). Mise en scène : Christine Zeppenfeld ; comédiennes : Valérie Le Louédec et Magali Bruneau ; conception multimédia : Alain Bonardi et Nathalie Dazin ; musique : Stéphane Grémaud ; lumières : Thierry Fratissier.

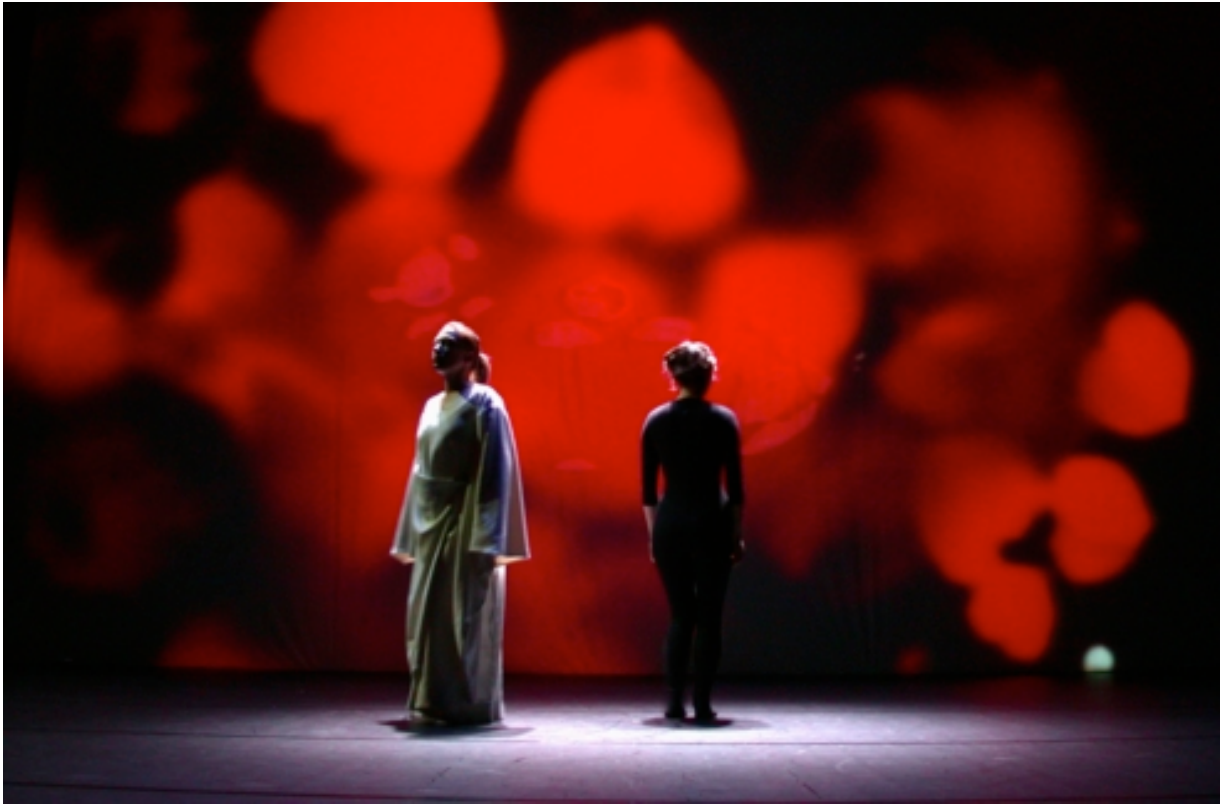

**Figure 1.** *Exemple de génération d'images sur l'écran de fond de scène dans* La traversée de la nuit *(Valérie Le Louédec à gauche, Magali Bruneau à droite; photographie : Julien Piedpremier).*

### 4.2. Description informatique du système homme-machine

L'implémentation informatique du système homme-machine est fondée sur un réseau de neurones d'analyse de la voix en entrée et un système multi-agents générateur d'images en sortie.

Le système informatique multimédia temps réel mis en œuvre est constitué en entrée d'un réseau de neurones destiné à reconnaître des états émotionnels dans la voix de la comédienne et en sortie d'un système multi-agents générateur d'images projetées sur l'écran. L'ensemble a été codé en utilisant la plateforme de développement graphique temps réel Max/MSP/Jitter.

Le réseau de neurones a été entraîné en mode supervisé pendant plusieurs mois par rapport à une liste d'états émotionnels que s'imposait la comédienne durant la lecture du texte complet. La voix en entrée est traitée phrase par phrase, chacune donnant lieu au calcul d'un vecteur de douze composantes : quatre d'entre elles concernent la prononciation des voyelles (formants), quatre d'entre elles représentent le caractère bruité de la voix et donc la prononciation des consonnes; les quatre derniers paramètres s'attachent à la prosodie (courbe d'amplitude de la voix dans la phrase). Pour chaque vecteur présenté en entrée, le réseau de neurones fournit un état émotionnel « reconnu ».

Le système multi-agents permet la génération temps réel d'images projetées en fond de scène. Les agents sont comme des « colleurs d'affiches » dynamiques qui construiraient ensemble des images toujours renouvelées.

− Chaque agent possède un petit modèle psychologique de sensibilité (positive ou négative), qui réagit selon les séquences de texte, aux états émotionnels du réseau de neurones. Il en résulte, en fonction de ce qu'indique le réseau de neurones, et en fonction des poids de sensibilité, une humeur qui conditionne leur « volonté » d'accomplir les tâches à mener.

− Les agents coopèrent à un but qui est l'optimisation d'une fonction d'utilité de l'image (une différente par séquence de texte).

- Les agents se coordonnent dans l'exécution de ce but commun par rapport à l'état émotionnel reconnu par le réseau de neurones, par un mécanisme de compensation d'humeur : ceux qui sont « d'excellente humeur » (grande valeur positive) concèdent un peu de leur ardeur à ceux qui ont une humeur très négative.
- Les agents communiquent entre eux deux à deux à période fixe en se transmettant leurs humeurs respectives.
- L'environnement des agents est constitué des états émotionnels reconnu par le réseau de neurones, du repère d'événement indiquant à quel endroit on se trouve dans la pièce et de valeurs propres à la séquence correspondante du texte et des indications d'un agent-observateur indiquant les qualités de l'image globale construite.

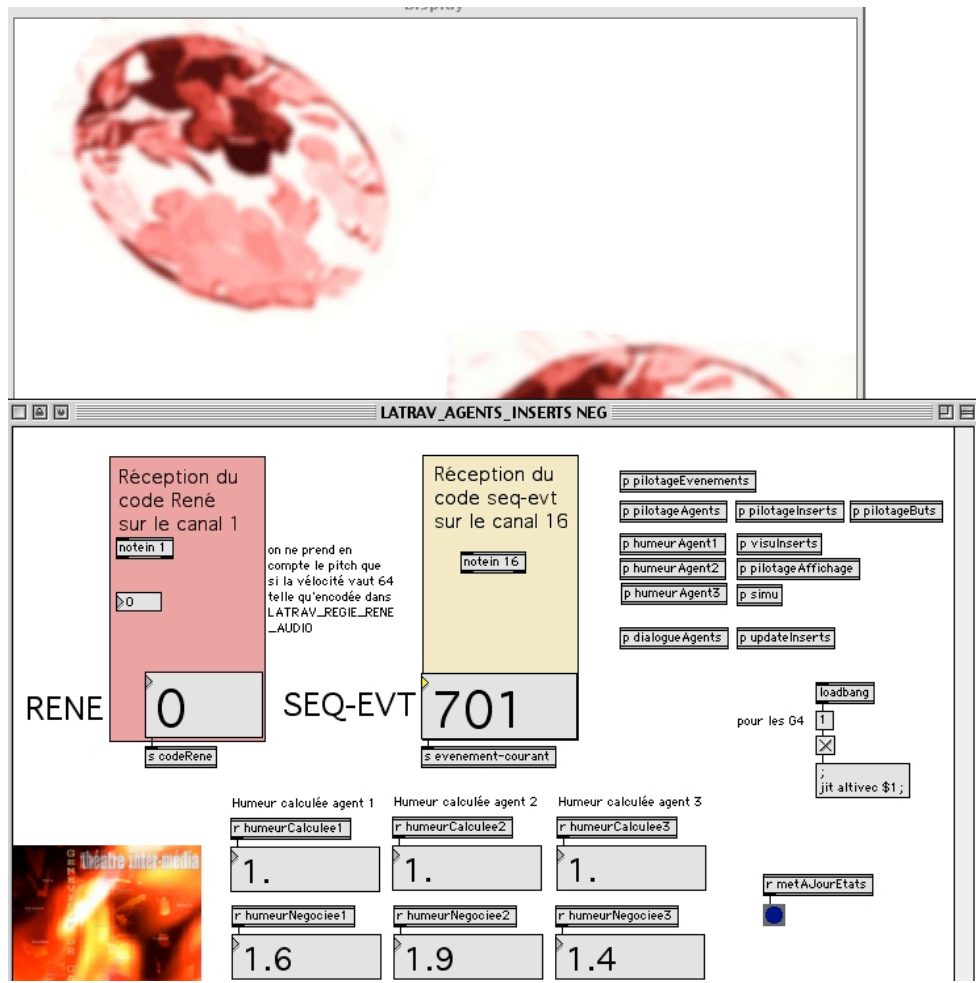

**Figure 2.** *Exemples de patchs Max/MSP/Jitter. En arrière, deux agents autonomes portant des fragments d'images dans La traversée de la nuit ; en avant, une partie de l'écran de pilotage (source : Alain Bonardi).*

## 5. CONCLUSION

Nous avons montré comment l'approche informatique de la similarité fondée sur la fouille de données interactive peut inspirer de nouvelles modalités de mise en scène de théâtre associant acteurs et ordinateurs.

Sortir de la mise en scène traditionnelle fondée sur les ontologies conduit irréversiblement à un affadissement de l'instanciation au profit de la manipulation active de contenus numériques passant par des transformations de données souvent irréversibles. En cela, mixer une compilation ou une séquence musicale dans un logiciel *ad hoc* ressemble en profondeur à l'établissement *live* face au public d'une continuité dramatique, lorsque les ordinateurs, devenus acteurs, et les comédiens, se provoquent mutuellement. Dans les deux situations, la machine est engagée dans un fonctionnement heuristique.

Remarquons enfin que ces nouvelles approches de théâtre inter-média ouvrent des possibilités de simulation homme-machine : le metteur en scène peut utiliser ces systèmes pour la simulation dynamique de ses idées.

## 6. RÉFÉRENCES BIBLIOGRAPHIQUES